\documentclass{article}

\usepackage{PRIMEarxiv}

\usepackage[utf8]{inputenc}
\usepackage[T1]{fontenc}
\usepackage[colorlinks=true, citecolor=blue, urlcolor=blue, linkcolor=blue]{hyperref}
\usepackage{url}
\usepackage{booktabs}
\usepackage{amsfonts}
\usepackage{nicefrac}
\usepackage{microtype}
\usepackage{fancyhdr}
\usepackage{graphicx}
\usepackage{amsmath,amssymb}
\usepackage{subcaption}
\usepackage{enumitem}
\usepackage{xcolor}
\usepackage{mdframed}
\graphicspath{{media/}}

\title{SR-TTT Does Not Learn Retrieval:\\ A Correction and Mechanistic Post-Mortem of\\ Surprisal-Aware Residual Test-Time Training
\thanks{\textit{\underline{Citation}}:
\textbf{Swamynathan V P. SR-TTT Does Not Learn Retrieval: A Correction and Mechanistic Post-Mortem of Surprisal-Aware Residual Test-Time Training. 2026.} This version (v2) supersedes and corrects arXiv:2603.06642v1.}
}

\author{
  Swamynathan V P \\
  Department of Computer Science and Engineering \\
  Amrita School of Computing \\
  Amrita Vishwa Vidyapeetham \\
  Chennai, Tamil Nadu, India \\
  \texttt{vpswamynathan0311@gmail.com} \\
}

\begin{document}
\maketitle

\begin{mdframed}[backgroundcolor=red!5, linecolor=red!60, linewidth=1pt]
\textbf{Correction notice.} Version 1 of this paper reported that SR-TTT improved Needle-in-a-Haystack exact match by $+23\%$ (depth $0.50$) and $+20\%$ (depth $0.75$) over a pure TTT baseline at 2048-token contexts. \textbf{These results were artifacts of evaluation and implementation flaws}, principally an off-by-one label alignment that reduced the task to copying an answer already present in the input, compounded by non-causal information flow in both the residual cache and the TTT inner loop. Under a corrected, leak-free protocol the reported advantage vanishes entirely (zero discordant trials out of 750 paired comparisons). All headline claims of v1 are retracted. This version documents the flaws, provides a corrected implementation with runtime causality verification, and reports a mechanistic post-mortem: even with the evaluation fixed, trainable retrieval machinery added, and storage guaranteed by an oracle, surprisal-gated exact-KV caching fails to learn needle retrieval, for reasons we measure and decompose.
\end{mdframed}

\begin{abstract}
Test-Time Training (TTT) language models replace the KV-cache with fast weights updated during inference, achieving $O(1)$ memory but suffering catastrophic failure on exact-recall tasks. Version 1 of this work proposed SR-TTT, which routes high-surprisal tokens to a sparse exact-attention Residual Cache, and reported large Needle-in-a-Haystack gains. We show those gains were evaluation artifacts: the loss and metric read logits \emph{at} the answer positions rather than one position earlier, training both models to copy an answer already visible in their input (a model trained on retrieval-\emph{impossible} data reaches 100\% accuracy under the flawed metric); additionally, the cache attended non-causally over future tokens, including the answer itself. We release a corrected implementation with startup causality self-tests, then ask whether the SR-TTT hypothesis survives correction. It does not, and the failure decomposes into two independent, separately measured bottlenecks. \textbf{Storage:} surprisal gating is systematically position-biased---the TTT reconstruction loss requires burn-in before a needle becomes \emph{relatively} surprising, so early-context needles are stored at near-zero rates (0--1\% containment at depth 0.1) exactly where long-context memory matters most. \textbf{Addressing:} with storage solved by an oracle and with new trainable read-time projections, per-slot attention supervision raises addressing mass $2.5\times$ (0.06$\to$0.15) yet token accuracy is statistically unchanged, and retrieval extracts only $\approx$0.06 nats of the 2.30-nat needle; position-free content addressing cannot resolve \emph{ordered} slots whose contents are near-interchangeable. Exact match remains 0\% in all $2{,}250$ paired trials across all corrected conditions. We retract the claims of v1 and offer the corrected codebase, diagnostic protocol, and negative results as a cautionary reference for surprise-gated memory architectures.
\vspace{1em}

\end{abstract}

\section{Introduction}

The premise of version 1 stands: fixed-size recurrent states, including TTT fast weights~\cite{sun2024ttt}, face a fundamental recall bottleneck once sequence length exceeds state capacity~\cite{arora2024based, arora2024zoology}, and rare tokens from early context are overwritten by subsequent updates~\cite{liu2024lost}. Our corrected experiments confirm this motivation cleanly: a pure TTT baseline, trained with a genuine language-modeling objective and evaluated without leaks, scores 0\% generation exact-match on needle retrieval at every context length and depth tested.

What does not stand is the proposed remedy---or, more precisely, the evidence that was offered for it. Version 1 reported that routing high-surprisal tokens to an exact-attention Residual Cache recovered $+23\%$ exact match at 2048 tokens. Prompted by anomalies in those results (retrieval that was \emph{hardest} at the shortest context, a stale chance line, and a sign flip at 1024 tokens that went undiscussed), we audited the training and evaluation code and found the result was manufactured by the pipeline rather than the mechanism. This paper does three things:

\begin{enumerate}[nosep]
    \item \textbf{Documents the flaws} (Section~\ref{sec:flaws}), including two ``smoking gun'' experiments: under the original metric, a model trained on data where retrieval is information-theoretically impossible reaches 100\% accuracy within 100 steps; and perturbing only the final 8 input tokens measurably changes logits at position 0 when the cache is enabled.
    \item \textbf{Releases a corrected implementation} (Section~\ref{sec:corrected}) with strict causality by construction, verified by self-tests that execute at startup, plus a paired evaluation protocol (identical seeded samples, generation-based exact match, Wilson intervals, exact McNemar tests).
    \item \textbf{Reports a mechanistic post-mortem} (Section~\ref{sec:results}). Beyond re-running the original comparison (which nulls), we give the mechanism its best chance: trainable read-time cache projections, per-slot oracle attention supervision, a needle-distance curriculum, and an oracle-storage upper-bound evaluation. Retrieval still fails, and the diagnostics localize the failure to two independent causes---position-biased storage and content-addressing that cannot resolve ordinal structure.
\end{enumerate}

We believe the corrected codebase, the diagnostic decomposition (storage $\to$ addressing $\to$ readout), and the specific failure modes are useful to the growing literature on surprise-gated memory~\cite{behrouz2025titans} precisely because they were measured where such architectures are rarely instrumented.

\section{Related Work}

\textbf{Test-Time Training.} TTT was introduced for distribution-shift robustness~\cite{sun2020ttt} and later formalized as a sequence-modeling primitive replacing the Transformer KV-cache~\cite{vaswani2017attention, sun2024ttt}, with subsequent efficiency improvements~\cite{zhang2025tttright}. The exact-recall limitation of compressed recurrent states that motivated v1 is confirmed, not contradicted, by our corrected results.

\textbf{Surprise-gated and hybrid memory.} Titans~\cite{behrouz2025titans} uses gradient-based ``surprise'' to gate memorization into a neural memory; for an MSE inner objective, reconstruction loss and its gradient magnitude are monotonically related, making SR-TTT's routing signal a close relative. Our storage-bottleneck finding (Section~\ref{sec:storage})---that reconstruction-loss surprise is position-biased because it requires burn-in---is therefore directly relevant to this family. Hybrid compressed-exact designs~\cite{arora2024based, dong2024less, jiang2024minference} and learned cache-retention policies~\cite{trimkv2025, zhang2024h2o, hooper2024kvquant, nawrot2024dynamic} address related trade-offs from the Transformer side; gated linear RNNs~\cite{gu2023mamba, peng2023rwkv, gdwm2026} from the recurrent side.

\textbf{Evaluation integrity.} Our flaw analysis adds to the catalogue of ways Needle-in-a-Haystack~\cite{kamradt2023niah} pipelines can silently degenerate: teacher-forced metrics that reward copying, per-model (unpaired) test sets, and underpowered cell counts. We adopt paired McNemar testing as the default for A/B retrieval comparisons.

\section{The Original Architecture (Summary)}
\label{sec:arch}

SR-TTT augments a 4-layer TTT-Linear backbone ($d_{model}{=}256$, 15.8M parameters, RoPE~\cite{su2021roformer}) with, per layer: a \emph{Surprisal Filter} flagging tokens whose inner-loop reconstruction loss $L_t = \lVert z_t - v_t \rVert^2$ exceeds an EMA-smoothed 95th-percentile threshold (with a chunk-level co-condition); a fixed-capacity \emph{Residual Cache} storing flagged tokens with priority eviction; and attention over the cache fused into the stream as $\text{Output} = \text{TTT}(x) + \alpha \cdot \text{CacheAttn}(x)$ with $\alpha = \mathrm{clamp}(\theta, 0, \alpha_{max})$. A two-stage curriculum (7{,}000 steps backbone-only, then 3{,}000 steps with the backbone frozen and the cache path trainable) was used throughout; this element of v1 survives unchanged.

\section{Flaws in the Version 1 Evaluation}
\label{sec:flaws}

\subsection{F1: Off-by-one label leak (fatal)}
Sequences end with \texttt{...\ Answer: <needle>}, and the code defined the query position as the index of the \emph{first answer token}. Both the training loss and the evaluation metric then read \texttt{logits[q : q+L]} against the L answer tokens. In next-token convention, the logit at position $q$ is produced \emph{after} the model has consumed answer token 0 as input: the target at every answer position is the token the model is currently reading. With tied embeddings this is a trivial identity map. The correct indexing is \texttt{logits[q-1 : q+L-1]}.

\textbf{Smoking gun.} We trained the original pure-TTT model on sequences in which the answer alphabet never appears anywhere in the context---retrieval is information-theoretically impossible---under the original indexing: token accuracy reaches \textbf{100\% within 100 steps}. Under corrected indexing, the identical setup remains at chance. The v1 training curves (90\%+ answer accuracy) measured this copy circuit. It also explains the strangest feature of the v1 heatmaps: accuracy of 11\%/80\%/0\% at 1024/2048/4096---genuine retrieval is \emph{easiest} at the shortest context, whereas a position-tuned copy circuit peaks exactly at the trained length.

\subsection{F2: Non-causal cache}
The v1 block inserted \emph{all} surprising tokens from the full sequence into the cache and then attended with \texttt{is\_causal=False}: every position could read the future, including the answer tokens' own cached entries (high-surprisal alphanumerics are precisely what the filter selects). Empirically, perturbing only the last 8 input tokens changed logits at position 0 by $0.21$ with the cache enabled. The $\alpha$ gates opening in Stage 2 therefore reflected exploitation of leaked future content, not needle rescue. The EMA threshold was additionally computed from a global quantile over the full sequence---a second future channel.

\subsection{F3: TTT within-window future leak}
Outputs used fast weights already updated on the entire 64-token window, so tokens conditioned on their own window's future (measured leak: $0.31$ within-window). The final window contains \texttt{Answer: <needle>}, so even the ``pure'' baseline saw the answer through this path.

\subsection{F4: Protocol and statistical problems}
(i) No language-modeling objective existed; only the (leaky) answer cross-entropy was trained, contradicting the described setup. (ii) Evaluation fed 64-token chunks in \emph{separate} forward passes while fast weights were re-initialized every forward: the baseline had no cross-chunk memory \emph{by construction}, making the comparison structurally rigged in favor of the cache. (iii) Exact match was teacher-forced argmax, not generation. (iv) $n{=}30$ trials per cell, with A and B evaluated on \emph{different} random samples; the headline $+23\%$ is 3/30 vs.\ 10/30 (Fisher exact $p\approx0.06$), and the 1024-token column favored the baseline by 6--8\%, which v1 did not discuss. (v) The plotted ``Random 25\%'' chance line was a stale artifact of an earlier 4-way design.

\section{Corrected Implementation and Upgraded Mechanism}
\label{sec:corrected}

\textbf{Corrections} (all verified by startup self-tests that assert bit-identical logits at unperturbed positions, cache on and off): corrected answer indexing everywhere; chunked-causal TTT (output computed from the state of \emph{previous} chunks only, 16-token update lag); per-window attend-\emph{then}-insert cache order with a strict position mask ($\text{pos}_{\text{cache}} < \text{pos}_{\text{query}}$) and sequential EMA thresholding; a genuine next-token LM objective in Stage 1; single full-sequence forward at evaluation (matching training); exact match measured by greedy \emph{generation}; paired evaluation on identical seeded samples with Wilson intervals and exact McNemar tests; honest chance lines. The TTT recurrence runs in fp32 with the learnable inner learning rate clamped inside its stability region ($\approx 1/d_{head}$ for layer-normed keys), eliminating a mixed-precision divergence we encountered at scale.

\textbf{Upgrades to give retrieval its best chance.} Because v1's cache stored keys produced by frozen projections---leaving \emph{no gradient path} by which addressing could improve---the corrected mechanism stores detached pre-TTT hidden states and computes cache keys/values/queries \emph{at read time} through new projections trained in Stage 2 ($\approx$1.05M parameters). We further add: (i) \emph{per-slot oracle attention supervision} (train-time only, ablated): the answer query for token $i$ is supervised toward the cache slot holding needle token $i$; (ii) a \emph{distance curriculum} annealing the needle$\to$query gap from 128 tokens to the sequence maximum; (iii) \emph{digit needles} (``The pass key is 3 8 2 9 1 7 4 6''), replacing rare BPE shards with clean single-digit tokens (chance = 10\%/token, prior entropy $\ln 10 \approx 2.303$ nats); (iv) train-time force-insertion of the needle into the cache (never at evaluation); and (v) \emph{mechanism diagnostics} logged throughout: \emph{containment} (fraction of needle tokens resident in the cache at query time), \emph{natural flagging} (would the filter have stored them unforced?), and \emph{addressing mass} (attention weight on the correct slot).

\section{Results}
\label{sec:results}

\textbf{Setup.} TinyStories~\cite{eldan2023tinystories} haystacks, GPT-2 tokenizer, 15.8M-parameter models trained at 2048 tokens on a single T4; Stage 1 trained once (final LM loss 3.46, perplexity $\approx$32; answer accuracy pinned at the 10\% digit prior, as it must be for a model with no retrieval mechanism) and forked into two Stage-2 variants: \emph{B-full} (oracle supervision + curriculum) and \emph{B-noOracle} (curriculum only). Evaluation: 50 trials/cell over lengths $\{1024, 2048, 4096\}$ $\times$ depths $\{0.1, 0.25, 0.5, 0.75, 0.9\}$, identical seeded samples for all models.

\subsection{The corrected comparison nulls completely}
Generation exact match is \textbf{0\% in every cell for every model}---baseline and both cache variants---yielding \emph{zero} discordant pairs in all 750 paired trials per comparison (2{,}250 total; pooled McNemar $p=1.0$ throughout). Token accuracy sits at or below the digit prior for all models (pooled means, B-full: 6.7\%/10.7\%/8.2\% at 1024/2048/4096; baseline: 3.0\%/9.7\%/4.3\%). The v1 effect does not merely shrink under correction; it is not detectably present.

\subsection{Bottleneck 1: surprisal storage is position-biased}
\label{sec:storage}
\begin{figure}[htbp]
    \centering
    \includegraphics[width=0.95\textwidth]{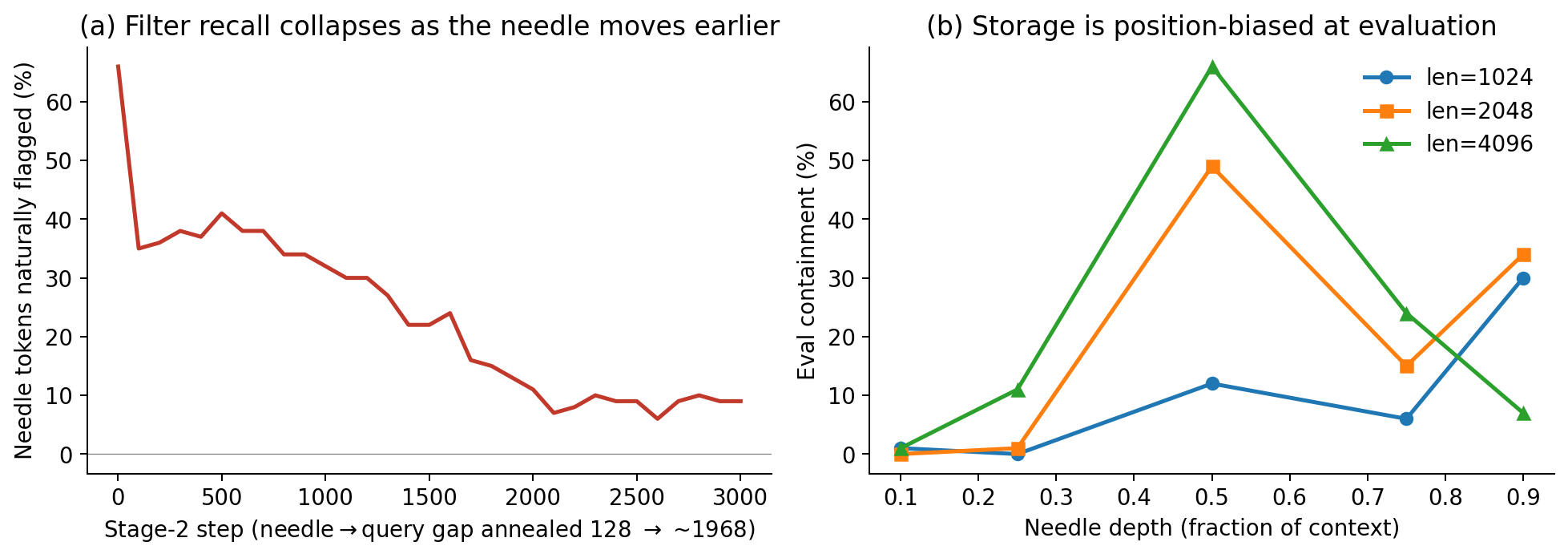}
    \caption{The storage bottleneck. (a) Natural (unforced) needle flagging during Stage-2 training collapses as the distance curriculum moves the needle earlier in context. (b) At evaluation, cache containment of the needle is near zero at shallow depths and peaks mid-sequence: reconstruction-loss surprise requires burn-in before a needle is \emph{relatively} surprising.}
    \label{fig:storage}
\end{figure}

Figure~\ref{fig:storage} shows the two views. During Stage-2 training, the fraction of needle tokens the filter would have flagged \emph{unforced} collapses from $\approx$38\% to 7--9\% as the curriculum moves the needle earlier in the sequence (a). At evaluation---where nothing is forced---containment is near zero for early needles (0--1\% at depth 0.1) and peaks mid-sequence (49--66\% at depth 0.5 for lengths 2048--4096) (b). The mechanism is a burn-in effect: early in a sequence the fast weights have adapted to nothing, so \emph{all} tokens have high reconstruction loss and the needle is not \emph{relatively} surprising; the EMA percentile threshold cannot separate it. Surprisal gating thus fails precisely for early-context needles---the case long-context memory exists to serve. Because gradient-magnitude ``surprise'' is monotone in reconstruction loss for MSE objectives, this bias plausibly extends to related surprise-gated designs~\cite{behrouz2025titans} and deserves direct measurement there.

\subsection{Bottleneck 2: addressing cannot resolve ordered slots, and better addressing does not become accuracy}
\label{sec:addressing}
\begin{figure}[htbp]
    \centering
    \includegraphics[width=0.6\textwidth]{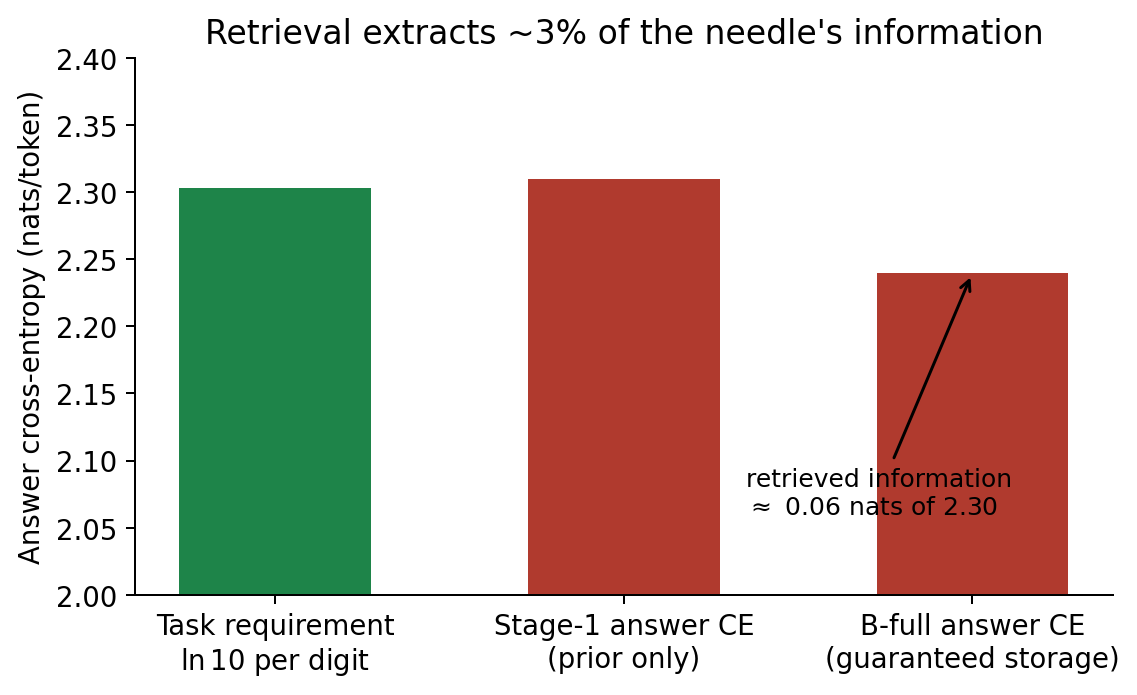}
    \caption{Information budget under oracle storage at the training length: retrieval recovers $\approx$0.06 nats/token of the 2.303-nat digit needle.}
    \label{fig:budget}
\end{figure}

To isolate addressing from storage, we re-evaluated both variants with \emph{oracle storage}: the needle force-inserted into every layer's cache at evaluation (30 trials/cell; containment 100\% everywhere by construction; an upper bound, not the method). Exact match remains \textbf{0/450 for both variants}. Token accuracy at the training length rises to 17.4\% (B-full) and 16.1\% (B-noOracle)---the 10\% prior plus roughly 7 points of genuine but far-too-weak retrieval; at $17\%$/token, an 8-token exact match has probability $\sim 10^{-6}$. In information terms (Figure~\ref{fig:budget}), the answer cross-entropy improves from the prior's 2.31 to only 2.24 nats/token: $\approx$0.06 of the needle's 2.30 nats is retrieved.

\begin{figure}[htbp]
    \centering
    \includegraphics[width=0.95\textwidth]{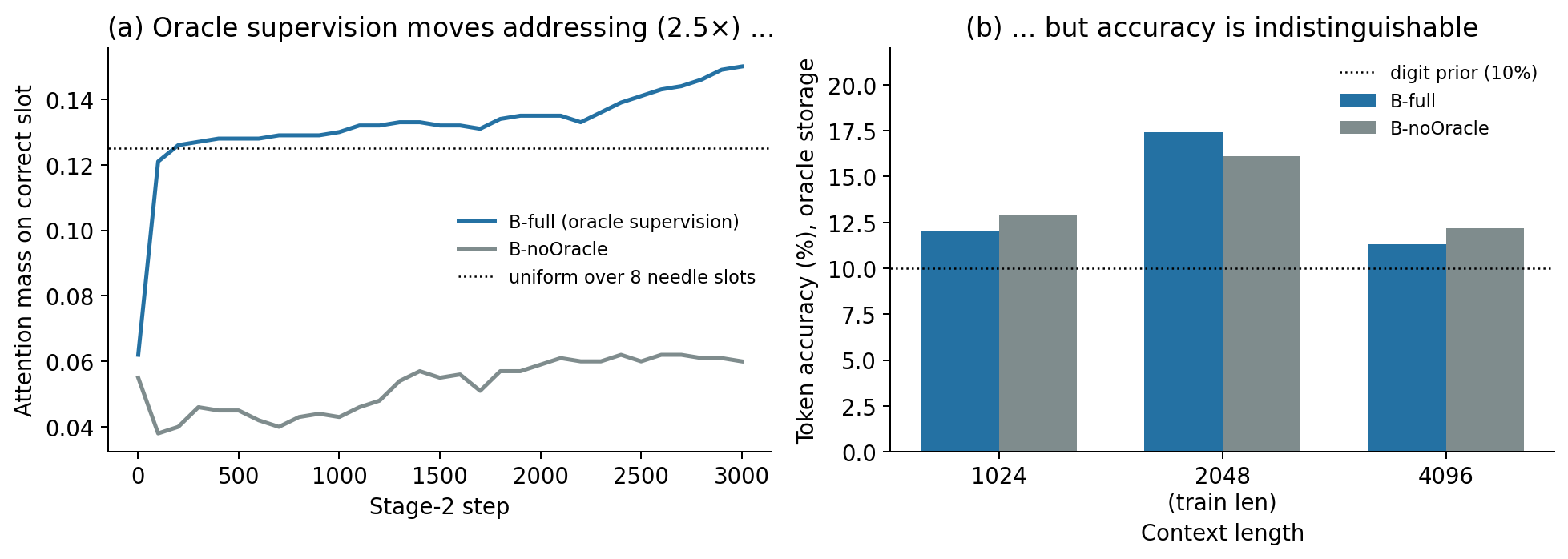}
    \caption{The addressing--accuracy dissociation under oracle storage (containment $=100\%$). (a) Per-slot oracle supervision raises addressing mass $2.5\times$ over the unsupervised variant. (b) Token accuracy is nonetheless statistically indistinguishable between variants and an order of magnitude short of exact-match viability.}
    \label{fig:dissociation}
\end{figure}

The ablation reveals a dissociation (Figure~\ref{fig:dissociation}): oracle supervision demonstrably achieved its target, raising attention mass on the correct slot $2.5\times$ (0.06 $\to$ 0.15, saturating just above uniform-over-needle-slots), yet token accuracy is statistically indistinguishable between the variants. Improving \emph{where attention lands} did not improve \emph{what comes out}. Two compounding causes: (i) at 0.15 mass, 85\% of the value readout is still noise from wrong slots; and (ii) the entire cache pathway enters the residual stream through $\alpha \approx 0.08$, attenuating the retrieved signal before the LM head sees it. Behind both sits a structural limit we introduced deliberately: position-free content addressing (adopted so retrieval would not depend on RoPE phase between needle and query, enabling length generalization) discards exactly the feature---position---that distinguishes eight near-interchangeable cached digit states from one another. Content addressing retrieves \emph{sets}; ordered retrieval requires positional or successor structure in the cached representations. Consistent with query-side RoPE features, all metrics peak at the training length 2048 and degrade at 1024 and 4096.

\begin{table}[htbp]
\centering
\caption{Corrected paired evaluation (50 trials/cell, pooled over depths; generation exact match; identical seeded samples). ``Oracle storage'' force-inserts the needle at evaluation (upper bound). Chance token accuracy is 10\%.}
\label{tab:main}
\begin{tabular}{llccc}
\toprule
Condition & Model & Exact match & Token acc.\ (1024/2048/4096) & Pooled McNemar vs.\ A \\
\midrule
Natural storage & A (pure TTT)   & 0\% & 3.0\% / 9.7\% / 4.3\% & --- \\
Natural storage & B-full         & 0\% & 6.7\% / 10.7\% / 8.2\% & $p=1.0$ (0 discordant) \\
Natural storage & B-noOracle     & 0\% & 4.6\% / 10.2\% / 7.1\% & $p=1.0$ (0 discordant) \\
\midrule
Oracle storage  & B-full         & 0\% & 12.0\% / 17.4\% / 11.3\% & --- \\
Oracle storage  & B-noOracle     & 0\% & 12.9\% / 16.1\% / 12.2\% & --- \\
\bottomrule
\end{tabular}
\end{table}

\section{Discussion}

\textbf{What is retracted, what is learned.} The v1 claims---that SR-TTT rescues exact recall, and that the opening $\alpha$ gates validated selective routing---are retracted in full. What replaces them is more specific than ``it does not work'': the failure factorizes. Storage fails for early-context needles because relative-surprisal signals need burn-in; addressing fails for ordered content because position-free content matching is information-theoretically incapable of resolving interchangeable slots; and readout attenuates whatever survives through a small fusion gate. Each layer was measured with its own instrument (natural-flagging rate, containment, addressing mass, answer-CE information budget), and each would need to be solved for an SR-TTT-style design to function.

\textbf{Design implications.} A viable successor would need (i) a storage policy that is position-debiased---e.g., surprisal normalized by a position-conditional baseline, or a learned insertion policy trained against retrieval outcomes~\cite{trimkv2025}; (ii) order-capable addressing---cached entries carrying positional or successor structure, or chained retrieval in which generating token $i$ conditions the query for token $i{+}1$ on retrieved context; and (iii) a readout path whose gain is not throttled by a scalar gate tuned for LM stability. We caution that each is a substantive research problem, not a patch.

\textbf{Methodological recommendations.} Retrieval claims in this regime should (a) verify causality mechanically (perturbation self-tests at startup), (b) measure exact match by generation, (c) evaluate paired on identical samples with McNemar tests, and (d) instrument the mechanism (storage/addressing/readout) rather than reporting only end-task accuracy, since end-task nulls are uninformative about \emph{which} component failed.

\section{Limitations}
All experiments use 15.8M-parameter models, a single training seed, a single (digit) needle format, and a T4 compute budget; the two-bottleneck decomposition is demonstrated in this regime and its quantitative details may shift at scale. The oracle-storage and oracle-supervision conditions are diagnostic scaffolds, not deployable mechanisms. Our negative result bounds this architecture and training recipe; it does not prove that no surprisal-gated exact-KV design can work, though Section~\ref{sec:results} constrains what a working one must overcome.

\section{Conclusion}
Version 1 of SR-TTT reported large exact-recall gains that we now show were produced by an off-by-one label leak and non-causal information flow, not by the proposed mechanism; those claims are retracted. Under a corrected, causality-verified protocol the mechanism confers no measurable benefit, and a best-case instrumented variant---trainable addressing, oracle supervision, guaranteed storage---retrieves $\approx$3\% of the needle's information. The corrected implementation, self-tests, diagnostic protocol, and both positive-control and negative results are available at \url{https://github.com/swamynathanvp/Surprisal-Aware-Residual-Test-Time-Training}. We hope the post-mortem is useful both as a caution on evaluation integrity in needle benchmarks and as measured evidence on where surprise-gated exact-memory designs break.


\end{document}